\documentclass[default,iicol]{sn-jnl}% Default with double column layout

\usepackage{graphicx}
\usepackage{array}
\usepackage{cleveref}
\usepackage{multirow}
\usepackage{subcaption}
\newcolumntype{C}[1]{>{\centering\let\newline\\\arraybackslash\hspace{0pt}}m{#1}}
\usepackage{amsmath}
\usepackage[displaymath, mathlines]{lineno}
\usepackage{hyperref,xcolor}
\usepackage{csquotes}
\usepackage{placeins}
\usepackage{soul}
\usepackage{cleveref}
\DeclareMathOperator*{\argmax}{argmax}
\jyear{2021}%

\theoremstyle{thmstyleone}%
\theoremstyle{thmstyletwo}%

\theoremstyle{thmstylethree}%

\begin{document}

\title[Article Title]{Pho(SC)-CTC - A Hybrid Approach Towards Zero-shot Word Image Recognition}


\author*[1]{\fnm{Ravi} \sur{Bhatt}}\email{2020aim1008@iitrpr.ac.in}
\equalcont{These authors contributed equally to this work.}

\author[1]{\fnm{Anuj} \sur{Rai}}\email{2019aim1003@iitrpr.ac.in}
\equalcont{These authors contributed equally to this work.}

\author[2]{\fnm{Sukalpa} \sur{Chanda}}\email{sukalpa@ieee.org}

\author[1]{\fnm{Narayanan C.} \sur{Krishnan}}\email{ckn@iitrpr.ac.in}

\affil*[1]{\orgdiv{Department of Computer Science \& Engineering}, \orgname{Indian Institute of Technology Ropar}, \orgaddress{\city{Rupnagar}, \postcode{140001}, \state{Punjab}, \country{India}}}

\affil[2]{\orgdiv{Department of Computer Science and Communication}, \orgname{ Østfold University College}, \orgaddress{\city{Halden}, \postcode{1757}, \country{Norway}}}


\abstract{Annotating words in a historical document image archive for word image recognition purpose demands time and skilled human resource (like historians, paleographers). In a real-life scenario, obtaining sample images for all possible words is also not feasible. However, Zero-shot learning methods could aptly be used to recognize unseen/out-of-lexicon words in such historical document images. Based on previous state-of-the-art method for zero-shot word recognition \enquote{Pho(SC)Net} \cite{AnujRai}, we propose a hybrid model based on the CTC framework (Pho(SC)-CTC) that takes advantage of the rich features learned by Pho(SC)Net followed by a \enquote{connectionist temporal classification} (CTC) framework to perform the final classification. Encouraging  results  were  obtained  on  two  publicly  available historical   document   datasets   and   one synthetic   handwritten dataset, which justifies the efficacy of Pho(SC)-CTC and Pho(SC)Net.}

\keywords{PHo(SC)Net , CTC , Zero-shot word recognition , Historical Documents , Zero-shot learning , Word recognition.}



\maketitle
\thispagestyle{plain}
\pagestyle{plain}
%\linenumbers
%\pagenumbering
\section{Introduction}\label{sec1}

Historical documents narrates condition about human societies in the past. Easy availability and usability of image acquisition devices have led to the digitization  and archival of such historical documents. Searching for important and relevant information from the large pool of images in those digital archives is a challenging task. Earlier, end-to-end transcription of the text using OCR was a popular way to achieve this goal. However, the performance of OCR often depends on character-segmentation accuracy, which is  error prone, specially in the context of cursive handwritten text. Moreover, end-users of such digital archives (historians, paleographers etc.) are often not interested in an end-to-end transcription of the text, rather they are interested in specific document pages where a query incident, place name, person name has been mentioned. To cater to this requirement, word-spotting and recognition techniques play an important role as they help directly in document indexing and retrieval. 

Deep learning models have been quite successful in many document analysis problems. Thus, it is a natural fit for word recognition in historical documents as well. However, training deep networks for this problem is a challenging task due to many reasons. The lack of a large corpus of word images, partly due to the changes in the appearance of characters and spellings of words over centuries, makes it difficult to train a deep model. The complexity is further compounded by the requirement of learning large number of word labels, using only a small set of samples. In addition, the historical documents exhibit many undesirable characteristics such as torn and blurred segments, unwanted noise and faded regions, handwritten annotations by historians and artefacts; all of them contributing to the difficulty of the task. 
\begin{figure}[h]
\centering
\includegraphics[width=7cm,height=6cm]{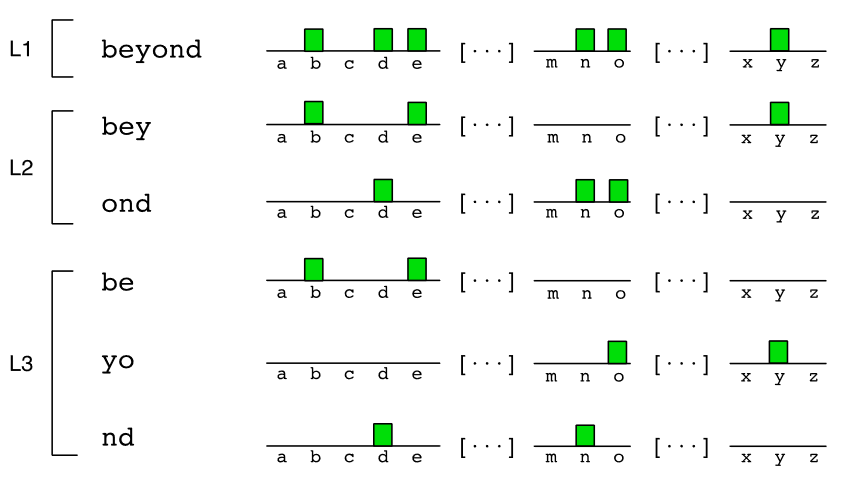}
\caption{PHOC histogram of a word at levels 1, 2, and 3. The final PHOC histogram is the concatenation of these partial histograms. \color{black}}
 \label{fig:phocExample}
\end{figure}
This work is focused on work recognition and extends it to the zero-shot learning (ZSL) setting.\color{black}

Classical word recognition involves training a machine learning model to recognise the words given the images containing them. It is assumed that the test query images also contain only the words that were presented during the training. However, in the ZSL setting, the test query images can contain words that the model did not see during training. This is a more challenging task requiring a visual representation (akin to the semantic embedding in ZSL literature) that can bridge the set of seen and unseen words.

In this work, we propose a machine learning model based on CTC framework that takes advantage of the rich features learned by Pho(SC)Net such that it can be used to recognise out-of-lexicon words. We call this model Pho(SC)-CTC. 
Overall we make the following contributions
\begin{itemize}
    \item We proposed a CTC-based framework for zero-shot word image recognition.
    \item We investigated the efficacy of Pho(SC)Net features while fusing them in a CTC framework for zero-shot word recognition. 
    \item The proposed hybrid model Pho(SC)-CTC \color{black} significantly outperforms the state-of-the-art method \enquote{Pho(SC)Net } for zero-shot word recognition accuracy on a real-world dataset and a synthetic dataset.
  
\end{itemize} 

\section{Related Work}
Word spotting and recognition have been well explored over the last 25 years, with a spurt in deep learning based solutions in the recent past \cite{sudholtF16,himanis2017,tomas2017,kang2018gcpr,chanda2018,krishnandas2018,dutta2018,alex2009,krishnan2019hwnet,prl_alicia_2020,wolf20}. The seminal work \cite{sudholtF16} on word spotting involved training of a CNN for predicting the PHOC representation \cite{almazanPami2014} (Figure \ref{fig:phocExample}). This work considered both contemporary as well as historical document images in their experiments. The proposed system can be used in both \enquote{Query By Example}(QBE) and \enquote{Query By String}(QBS) settings. In \cite{krishnandas2018}, the authors proposed an End2End embedding scheme to learn a common representation for word images and its labels, whereas in \cite{dutta2018}, a CNN-RNN hybrid model was proposed for handwritten word recognition. The method proposed in \cite{himanis2017} uses a deep recurrent neural network (RNN) and a statistical character language model to attain high accuracy for word spotting and indexing. A recent work on robust learning of embeddings presents a generic deep CNN learning framework that includes pre-training with a large synthetic corpus and augmentation methods to generate real document images, achieving state-of-the-art performance for word spotting on handwritten and historical document images \cite{krishnan2019hwnet}. In \cite{KRISHNAN2020}, the authors introduce a novel semantic representation for word images that uses linguistic and contextual information of words to perform semantic word spotting on Latin and Indic scripts. From the brief discussion it is evident that though deep learning based methods have been used in the recent past for word spotting and recognition tasks, zero-shot word recognition - recognizing a word image without having seen examples of the word during training; has not been studied. Only \cite{chanda2018} explore Latin script word recognition problem in the ZSL framework; however the number of test classes were limited and no publicly available datasets were used in their experiments.

ZSL techniques have demonstrated impressive performances in the field of object recognition/detection  \cite{Li_2018_CVPR,Zhang_2018_CVPR,Annadani_2018_CVPR,Niu_2018_CVPR,Li2019,Xie_2019_CVPR}. Li et.al. \cite{Li_2018_CVPR} propose an end-to-end model capable of learning latent discriminative features jointly in visual and semantic space. They note that user-defined signature attributes loose its discriminativeness in classification because they are not exhaustive even though they are semantically descriptive. Zhang et. al. \cite{Zhang_2018_CVPR}, use a Graph Convolutional Network (GCN) along with semantic embeddings and the categorical relationships to train the classifiers. This approach takes as input semantic embeddings for each node (representing the visual characteristic). It predicts the visual classifier for each category after undergoing  a series of graph convolutions. During training, the visual classifiers for a few categories are used for learning the GCN parameters. During the test phase, these filters are used to predict the visual classifiers of unseen categories \cite{Zhang_2018_CVPR}. In \cite{Annadani_2018_CVPR}, the objective functions were customized to preserve the   relations between the labels in the embedding space. In \cite{akatapami2016}, attribute label embedding methods for zero-shot and few-shot learning systems were investigated. Later, a benchmark and systematical evaluation of zero-shot learning w.r.t. three aspects, i.e. methods, datasets and evaluation protocol was done in \cite{akatacvpr2017}. In \cite{paulcvpr19}, the authors propose a conditional generative model to learn latent representations of unseen classes using the generator trained for the seen classes. The synthetically generated features are used to train a classifier for predicting the labels of images from the unseen object category. On similar lines \cite{Shi2017AnET} proposes a novel end to end trainable neural network architecture for scene text recognition, 
\cite{Long2020SceneTD} is a survey that summarizes the progress in scene text detection and recognition in the deep learning era.  \color{black}\newline
\begin{figure}[ht]
\centering
\includegraphics[width=8cm]{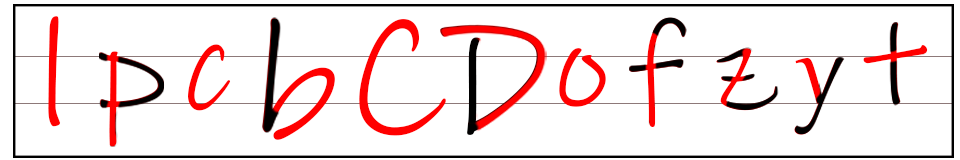}
\caption{11 primary shape attributes: ascender, descender, left small semi-circle, right small semi-circle, left large semi-circle, right large semi-circle, circle, vertical line, diagonal line, diagonal line at a slope of 135 degrees, and horizontal line}
 \label{fig:shapes}
\end{figure}
In summary, the current ZSL methods are mainly focused towards object detection. One must note that semantic attribute space in ZSL-based object detection is rich as attributes like colour and texture pattern play a crucial role. But in case of ZSL-based  word recognition the semantic attribute space is rather constrained due to the absence of such rich visual features. Further, the major bottleneck for ZSL based word recognition is the absence of the semantic embedding or attribute signature that establishes the relationship between the various word labels. This is further challenged by large number of word classes with relatively few examples.\newline
Graves et.al. \cite{ctc2006} introduces connectionist temporal classification, which is a framework that can be used to train sequential models and decode their output. The paper shows that CTC outperformed the baseline HMM model on speech recognition task. This motivates us to use CTC on handwritten word recognition as recognising the character sequence in the goal for both the approaches. The ZSL techniques discussed depend on the semantic embedding, CTC on the other hand is free from these dependencies.
In this work, we propose a novel CTC-based method which takes into account a rich feature representation method based for zero-shot word recognition. We hypothesize that the CTC model will be able to recognize the unseen words without the need for the visual description of the characters in the word. Further prior knowledge about the visual description can be successfully transferred to the CTC framework for improving its performance.
% \begin{figure}[ht]
% \centering
% \includegraphics[width=8cm]{PrimaryShapes.png}
% \caption{11 primary shape attributes: ascender, descender, left small semi-circle, right small semi-circle, left large semi-circle, right large semi-circle, circle, vertical line, diagonal line, diagonal line at a slope of 135 degrees, and horizontal line}
%  \label{fig:shapes}
% \end{figure}
\begin{figure*}[tp]
\centering
\includegraphics[width=15cm]{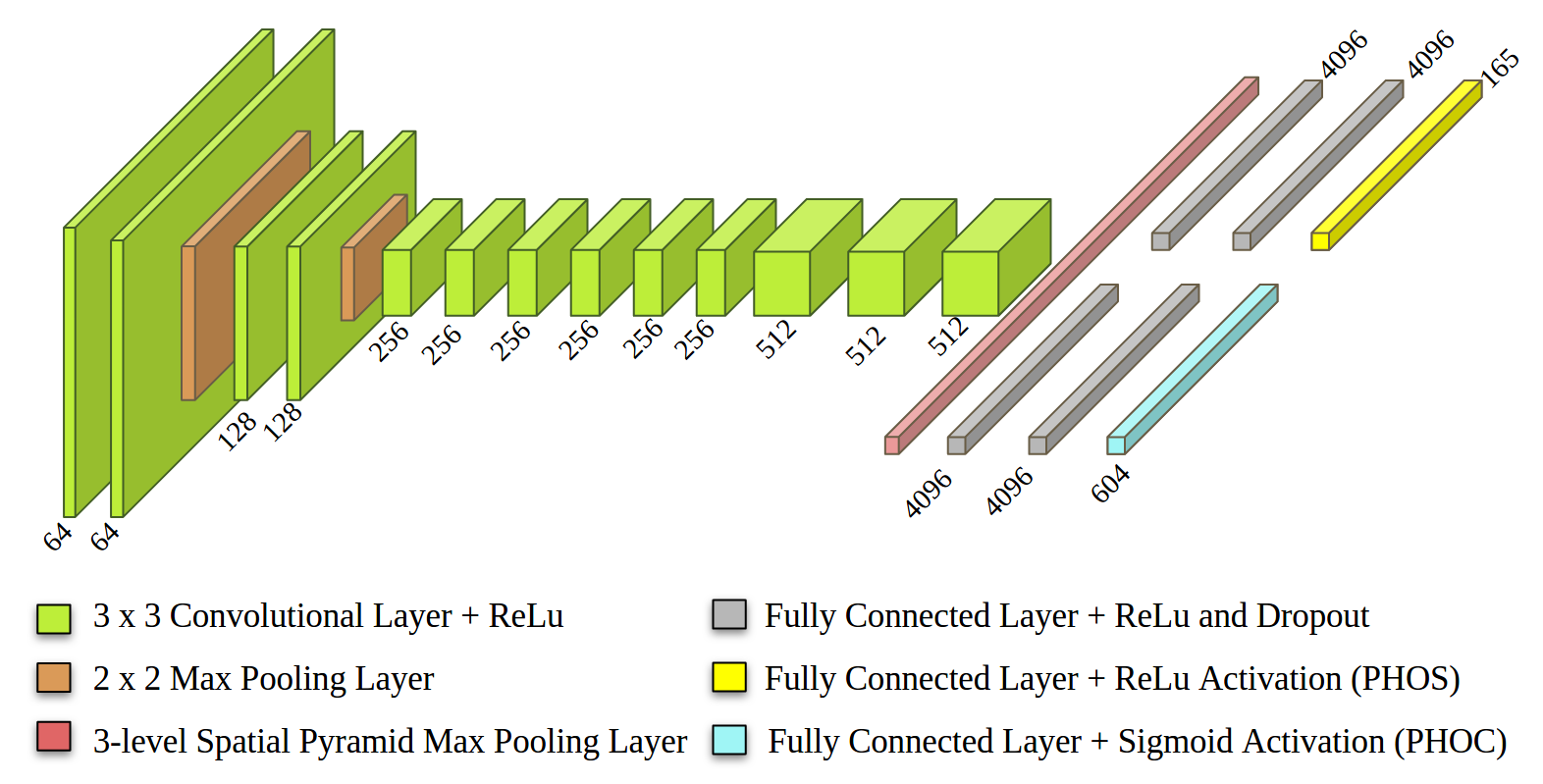}
\caption{Architecture of the multi-task Pho(SC)Net}
 \label{fig:combmodel}
\end{figure*}

\section{Methodology}
% \subsection{Problem Definition}
We begin by defining our problem of interest. Let $\mathcal{S} = \{ (x_i, y_i, c(y_i))\}_{i=1}^N$, where $x_i \in \mathcal{X}, y_i \in \mathcal{Y}^s, c(y_i) \in \mathcal{C}$, $\mathcal{S}$ stands for the training examples of seen word labels, $x_i$ is the image of the word, $y_i$ is the corresponding word label in $\mathcal{Y}^s = \{s_1, s_2, \ldots, s_K\}$ consisting of $K$ discrete seen word labels, and $c(y_i) \in \mathcal{R}^Q$ is the unique word label embedding or attribute signature that models the visual relationship between the word labels. In addition, we have a disjoint word label set $\mathcal{Y}^U = \{u_1, \ldots, u_L\}$ of unseen labels, whose attribute signature set $U = \{u_l,c(u_l)\}_{l=1}^L, c(u_l) \in \mathcal{C}$ is available, but the corresponding images are missing. Given $\mathcal{S}$ and $\mathcal{U}$, the task of zero-shot word recognition is to learn a classifier $f_{zsl}: \mathcal{X} \rightarrow \mathcal{Y}^u$ and in the generalized zero-shot word recognition, the objective is to learn the classifier $f_{gzsl}: \mathcal{X} \rightarrow \mathcal{Y}^u \cup \mathcal{Y}^s$. In the absence of training images from the unseen word labels, it is difficult to directly learn $f_{zsl}$ and $f_{gzsl}$. Instead, we learn a mapping ($\phi$) from the input image space $\mathcal{X}$ to the attribute signature space $\mathcal{C}$ that is shared between $\mathcal{Y}^s$ and $\mathcal{Y}^u$. The word label for the test image $x$ is obtained by performing a nearest neighbor search in the attribute signature space using $\phi(x)$. Thus, the critical features of zero-shot word recognition are the attribute signature space $\mathcal{C}$ that acts as a bridge between the seen and unseen word labels and the mapping $\phi$. In this work, we propose a novel attribute signature representation that can effectively model the visual similarity between seen and unseen word labels. The mapping $\phi$ is modeled as a deep neural network.
A very popular word label representation that can serve as the attribute signature for our problem is the pyramidal histogram of characters (PHOC). A PHOC is a pyramidal binary vector that contains information about the occurrence of characters in a segment of the word. It encodes the presence of a character in a certain split of the string representation of the word. The splits of different lengths result in the pyramidal representation. The PHOC allows to transfer information about the attributes of words present in the training set to the test set as long as all attributes in the test set are also present in the training set.  However, this constraint may be violated in the context of zero-shot word recognition. Further, the PHOC also misses the visual shape features of the characters as they appear in a word image. We also observe these limitations of PHOC from our experiments on unseen word recognition.  Literature also suggests parity between various representations that only encode the occurrence and position of characters within a word\cite{embedeval17}. Thus we are motivated to present a novel attribute signature representation that complements the existing word label characterizations.

\subsection{The Pyramidal Histogram of Shapes}
We propose the pyramidal histogram of shapes (PHOS) as a robust bridge between seen and unseen words. Central to the PHOS representation is the assumption that every character can be described in terms of a set of primitive shape attributes \cite{chanda2018}. We consider the following set of primitive shapes :-  ascender, descender, left small semi-circle, right small semi-circle, left large semi-circle, right large semi-circle, circle, vertical line, diagonal line, diagonal line at a slope of 135 degrees, and horizontal line. These shapes are illustrated in Figure \ref{fig:shapes}. Only the counts of these shapes is insufficient to adequately characterize each word uniquely. Inspired by the pyramidal capture of occurrence and position of characters in a word, we propose the pyramidal histogram of shapes that helps in characterizing each word uniquely.

\begin{figure*}[ht]
\centering
\includegraphics[width=15cm]{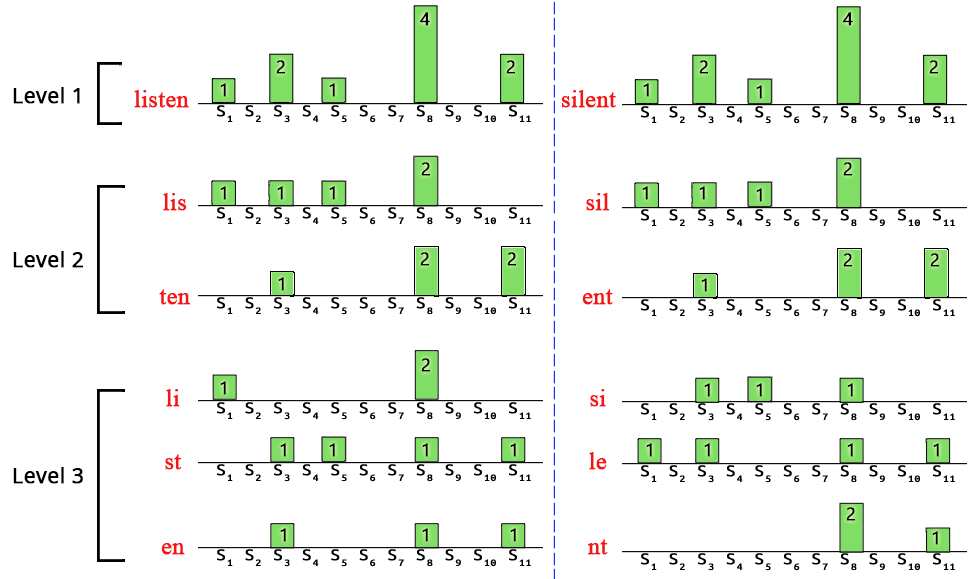}
\caption{Pyramidal structure of PHOS representation}
 \label{fig:pyramid}
\end{figure*}

The process of capturing the PHOS representation for a word is illustrated in Figure \ref{fig:pyramid}. At the highest level of the pyramid, there exists only a single segment, which is the entire word. Following the lines of \cite{almazanPami2014}, \color{black} at every level $h$ of the pyramid, we divide the word into $h$ equal segments thus giving equal weightage to each part. \color{black} Further, at every level $h$, we count the occurrence of the 11 primary shapes in every $h$ segments of the word.  The concatenation of the count vectors for every segment in a level and across all the levels of the pyramid results in the PHOS representation of the word. In this work, we have used levels 1 through 5, resulting in a PHOS vector of length (1+2+3+4+5)*11 = 165. Thus the PHOS vector encodes the occurrence and relative position of the shapes in the word string.

For example, let us consider a pair of anagrams \enquote{listen} and \enquote{silent} for 3 levels of segmentation. The segments at three levels for \enquote{listen} are: $L_1 = \{listen\}, L_2=\{lis,ten\}, L_3=\{li,st,en\}$. Similarly for \enquote{silent}:  $L_1 = \{silent\}$, $L_2=\{sil,ent\}$, $L_3=\{si,le,nt\}.$ The corresponding shape counts and their PHOS vector at each level for both words has been illustrated in Figure \ref{fig:pyramid} \color{black}.

For the zero-shot word recognition problem, it is important to encode the occurrence and relative position of characters within a word, as well as that of visual shapes. Therefore, we propose to use the concatenated PHOC and PHOS vector of a word as its attribute signature representation $\mathcal{C}$. Thus, the attribute signature representation for the word label $y_i$ is $[c_c(y_i),c_s(y_i)]$, where $c_c(y_i)$ and $c_s(y_i)$ are the PHOC and PHOS representations respectively.

\subsection{Pho(SC)Net Architecture}
Having defined the augmented attribute signature space $\mathcal{C}$, our next objective is to learn the mapping $\phi$ to transform an input word image into its corresponding attribute signature representation - PHOC+PHOS vector. We use the architecture of SPP-PhocNet\cite{sudholtF16} as the backbone for the Pho(SC)Net ($\phi$) that is used to predict the combined representation. The Pho(SC)Net is a multi-task network with shared feature extraction layers between the two tasks (PHOC and PHOS). The shared feature extraction network is a series of convolution layers, followed by a spatial pyramid pooling (SPP) layer. The SPP layer facilitates the extraction of features across multiple image resolutions. The Pho(SC)Net separates out into two branches after the SPP layer to output the two representations. The two branches contain two independent fully connected layers. As the PHOC representation is a binary vector, the PHOC branch ends with a sigmoid activation layer. On the other hand the PHOS representation being a non-negative vector, the PHOS branch ends with a ReLU activation layer. The multi-task Pho(SC)Net architecture is illustrated in Figure \ref{fig:combmodel}.

The output of the Pho(SC)Net for an input word image is the vector $\phi(x)=[\phi_{C}(x), \phi_{S}(x)]$, where $\phi_{C}(x)$ and $\phi_{S}(x)$ are the predicted PHOC and PHOS representations respectively. Given a mini batch of $B$ instances from the training set consisting of seen word images and their labels, we minimize the following loss function during training.
\begin{equation}
    L = \sum_{i=1}^B \lambda_c L_c(\phi_c(x_i), c_c(y_i)) + \lambda_s L_s(\phi_s(x_i), c_s(y_i))
\end{equation}
where $ L_c(\phi_c(x_i), c_c(y_i)) $ is the cross entropy loss between the predicted and actual PHOC representations, $L_s(\phi_s(x_i), c_s(y_i))$ is the squared loss between the predicted and actual PHOS representations, and $\lambda_c, \lambda_s$ are hyper-parameters used to balance the contribution of the two loss functions.

Given a test image $x$, the Pho(SC)Net is used to predict the PHOC and PHOS representations to obtain the predicted attribute signature representation $[\phi_c(x), \phi_s(x)]$. The word whose attribute signature representation has the highest similarity (measured as cosine similarity, Euclidean distance gave poor performance, see Tables [\ref{table:convzslresult},\ref{table:gzslresults},\ref{table:gzslresultseuclid}]\color{black}) with $[\phi_c(x), \phi_s(x)]$ is the predicted word label for the test image, in the conventional ZSL setting, as defined below
\begin{linenomath}
\begin{equation}
    \hat{y} = \argmax_{k\in\mathcal{Y}^U} \cos([\phi_c(x), \phi_s(x)]^T [c_c(k),c_s(k)])
\end{equation}
\end{linenomath}

\begin{figure}[ht]
\centering
\includegraphics[width=7cm,height=5cm]{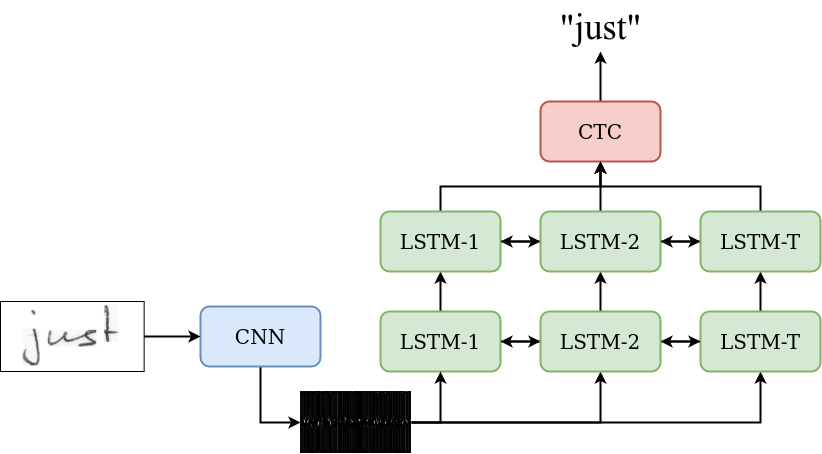}
\caption{Overview of CTC based neural network for handwritten word recognition}
 \label{fig:ctc_arch}
\end{figure}

\subsection{Connectionist temporal classification (CTC)}
Given a dataset of images containing words and the word label, it is difficult and time consuming to annotate each image. Fortunately, the Connectionist temporal classification framework, can train neural networks with sequential input and decode the output. In the case of handwritten word recognition, each character in the word does not need to be aligned to its exact location in the image.\newline
For a given input image $X$ and label $Y$, there can be various alignments possible between the input and the output. CTC \color{black} sums up the probabilities of all such alignments to calculate the likelihood $p(Y \mid X)$. Suppose we use the trivial approach of collapsing consecutive similar characters to one character to obtain the label from the decoded output. In that case, it becomes impossible to predict words that have consecutive duplicate characters. To solve this problem, CTC introduces a unique token called the blank token. If the output has two consecutive identical characters, the decoded output must have a blank character between the two characters. For example, consider the following outputs : 1) $AAAB = AB$ and 2) $AA - AB = AAB$ \newline
In the first example multiple instance of a character are considered to be a single character by the collapsing function, while in second example as two A's are separated by a '-'(special character in this case) they are considered to be two different characters. \newline
The easiest way to decode the RNN output is to take the most probable character at each time step, the best path decoding algorithm. There are two more algorithms to decode the RNN - output beam search and word beam search. In beam search decoding, text candidates (beams) are created and scored iteratively. An empty beam and a corresponding score are added to a list of beams at the start. The algorithm then iterates through the output matrix's all-time steps. Only the highest-scoring $k$ beams from the previous time-step are retained at each time-step. Here $k$ is the beamwidth. Word beam search makes use of a language model; therefore, using it for zero-shot word recognition will not be useful.\newline
The decoded output is further passed into a collapsing function as a single character could be encoded by duplicate characters and the special character. The decoding function removes duplicate characters and the special character and returns the predicted label.
\subsection{CTC Model \color{black} Architecture}
%\subsection{ \protect \st{ Pho(SC)-CTC} \color{blue} CTC Model \color{black} Architecture}
Our CTC model has two primary components, first is a CNN, followed by a RNN (2 Bi-directional Stacked LSTM cells). In this hybrid model, we considered the CNN part of the Phos(SC)Net, i.e. we feed the RNN part of the CTC model with the output of the last Convolution layer (which is the input to the first dense layer in the  PhoscNet architecture) (\textbf{Figure \ref{fig:combmodel}}). Max pooling layers are added such that the output of the CNN can be passed to the RNN.
The Pho(SC)-CTC architecture was trained using the following two different approaches.
\begin{itemize}
    \item \textbf{Pho(SC)-CTC\_Pre-Trained-Weights}: The weights of the convolutional layers of the CTC Model \color{black} are initialized with the trained weights of Pho(SC)Net. Henceforth, we will refer to this approach as {Pho(SC)-CTC} \color{black}
    
    \item \textbf{Vanilla-CTC} \color{black}: All weights in the entire network are obtained after training the whole end-to-end network from scratch.
\end{itemize}
Given a mini-batch of B instances from the training set consisting of seen word images and their labels, we minimize the following loss function during training.
\begin{linenomath}
\begin{equation}
    L = -\sum_{i=1}^{B} \ln(p(y_i \mid x_i)
\end{equation}
\end{linenomath}
Where $p(y_i \mid x_i)$ is the probability of getting $y_i$ as output given $x_i$ as input.

\begin{table*}[t]
\centering
\resizebox{\textwidth}{!}{%
\begin{tabular}{|c|c|c|c|c|}
\hline
\textbf{Split} & \textbf{Train Set } & \textbf{Validation Set } & \textbf{Test (Seen Classes)} & \textbf{Test (Unseen Classes)} \\ \hline
\multicolumn{5}{|c|}{\textbf{MFU Dataset}}                      \\ \hline
\textbf{MFU 2000}            & 36000(2000)  & 3600  & 4000  & 8000(1000)  \\ \hline
\multicolumn{5}{|c|}{\textbf{George Washington (GW Dataset)}} \\ \hline
\textbf{Split 1}             &   1585(374)     &    662(147)   &    628(155)   &    121(110)   \\
\textbf{Split 2}             &      1657(442)  &    637(165)   &    699(155)   &    114(100)   \\
\textbf{Split 3}             &   1634(453)     &    709(164)   &    667(148)   &    105(89)   \\
\textbf{Split 4}             &      1562(396)  &    668(149)   &    697(163)   &    188(152)   \\ \hline
\multicolumn{5}{|c|}{\textbf{IAM Handwriting Dataset}}        \\ \hline
\textbf{ZSL Split}       &   30414(7898)     &  2500(1326)     &    1108(748)   &   538(509)    \\ 
\textbf{Standard Split}  &   34549(5073)     &  9066(1499)     &    8318(1355)   &  1341(1071)     \\ \hline
\end{tabular}%
}
\caption{Details of dataset used for experiments \textit{Numbers inside the parentheses represent the number of word classes in the set.}}
\label{table:datasetsplit}

\end{table*}

\begin{table*}[h]
\centering
%\resizebox{\textwidth}{!}{%
\begin{tabular}{|l|c|c|c|c|c|c|c|}
\hline
\hline
\textbf{Split} & \multicolumn{3}{c|}{\textbf{Cosine Similarity}} & \multicolumn{3}{c|}{\textbf{Euclidean Distance}}\\
\hline
& \textbf{PHOC} & \textbf{PHOS} & \textbf{Pho(SC)} & \textbf{PHOC} & \textbf{PHOS} & \textbf{Pho(SC)}\\
%\multicolumn{1}{|C{3cm}|}{\textbf{Split}} & \multicolumn{1}{|C{3cm}|}{\textbf{PHOC}} & \multicolumn{1}{|C{3cm}|}{\textbf{PHOS}} & \textbf{Pho(SC)} 
\hline
\hline
\multicolumn{7}{|l|}{\textbf{MFU Dataset}} \\
\hline
MFU 2000 & .94 & .96 & \textbf{.98} & .90 &  .92 & .91\\ 
\hline
\hline
\multicolumn{7}{|l|}{\textbf{GW Dataset}} \\
\hline
GW Split 1           & .46 & .61 & \textbf{.68}  & .35 & .52 & .44\\
\hline
GW Split 2           & .64 & .72 & \textbf{.79}  & .54 &  .69 & .58\\
\hline
GW Split 3           & .65 & .71 & \textbf{.80} & .5 & .65 &  .66\\
\hline
GW Split 4           & .35 & \textbf{.62} &  .60 & .28 &  .4 &  .36\\ 
\hline
\hline
\multicolumn{7}{|l|}{\textbf{IAM Handwriting Dataset}} \\
\hline
ZSL Split            & .78 & .79 & .86 & .83 & .87 &  \textbf{0.9}\\
\hline
Standard Split       & .89 & .88 &  \textbf{.93} & .69 &  .76 &   .76\\ 
\hline
\hline
\end{tabular}%
% }
\caption{ZSL Accuracy on all the splits.}
\label{table:convzslresult}
\end{table*}

\section{Experiments}
\subsection{Datasets}
We validate the effectiveness of the Pho(SC) representation for the zero-shot word recognition problem on the following three datasets. 
\subsubsection{Most Frequently Used Words (MFU) Dataset}
A synthetic dataset was created from the most frequently used English words list as determined by n-gram frequency analysis of the Google's Trillion Word Corpus\footnote{\url{https://github.com/first20hours/google-10000-english}}. \color{black} $\mathcal{Y}^s$ was chosen to be the first 2000 words and the subsequent 1000 words were made part of $\mathcal{Y}^u$. Eight handwriting fonts were used to generate a total of 16000-word images (split into 12000 for training and 4000 for testing) for $\mathcal{Y}^s$ and 8000 for $\mathcal{Y}^u$. We ensure that the word labels of the 4000 test images of $\mathcal{Y}^s$ are present in the corresponding training set. These details are summarized in Table \ref{table:datasetsplit}. 

\subsubsection{George Washington (GW) Dataset}
The George Washington dataset\cite{gwdataset} exhibits homogeneous writing style and contains 4894 images of 1471 word labels.  We used the lower-case word images from the standard four-fold cross-validation splits accompanying the dataset to evaluate the Pho(SC)Net. We modified the validation and test sets to suit the zero-shot word recognition problem. Specifically, the test sets in each split was further divided into two parts: seen and unseen word label images, and the validation set contained only seen word label images. The details for each set across all the splits are presented in Table \ref{table:datasetsplit}.

\subsubsection{IAM Handwriting (IAM) Dataset}
IAM handwriting dataset\cite{iam_dataset_marti2002} is a multi-writer dataset that consists of 115320 word-images, from 657 different writers. We used the lowercase word-class images from this dataset to create train, validation, and test sets. Specifically we created two different splits. In the first split (overlapping writers ZSL split), we ensured that the writers in the test set are also part of the training set. Further, the test set contained both unseen and seen word labels, while the train and validation sets contained only seen word images. In the second split (standard ZSL split) derived from the standard split accompanying the dataset we removed the unseen word images from the validation set, and divided the test split further into seen word and unseen word images. The standard ZSL split has an additional challenge as the writers in the training, validation, and test sets are non-overlapping. The number of images and the (seen and unseen) word labels for each split is presented in Table \ref{table:datasetsplit}.

\begin{figure}[ht]
\centering
\includegraphics[width=8cm]{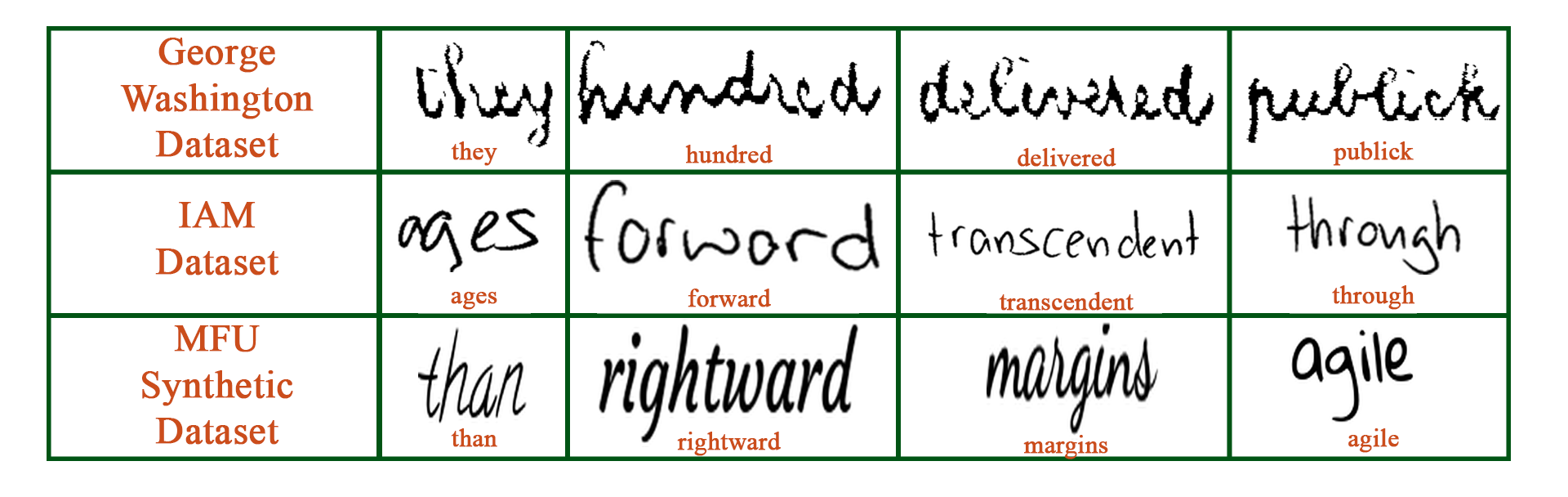}
\caption{Examples of word images and their labels from the three datasets}
 \label{fig:samples}
\end{figure}

The training set of all the three datasets were further augmented through shearing, and addition of Gaussian noise. Shearing range was chosen to be 20 and the deviation for the noise was calculated using a random number and a variability factor, higher the variability value higher is the deviation. \color{black} The size of a word image (in the training, validation, and test sets) depends on the length of the word and the handwriting style (font), but we needed images of uniform size for training. Hence, similar to \cite{krishnan2019hwnet}  \color{black} the binarized images were resized to the best fitting sizes (without changing the aspect ratio) and then padded with white pixels to get images of size 250*50. Figure \ref{fig:samples} presents a few examples of word images from the three datasets.

\begin{table*}[h]
\centering
\resizebox{\textwidth}{!}{%
\begin{tabular}{|l|c|c|c|c|c|c|c|c|c|}
\hline
\hline
\multicolumn{1}{|c|}{\textbf{Split}} &
  \multicolumn{3}{|c|}{\textbf{PHOC}} &
  \multicolumn{3}{|c|}{\textbf{PHOS}} &
  \multicolumn{3}{|c|}{\textbf{Pho(SC)}} \\ 
  \hline
\multicolumn{1}{|c|}{} &
  \multicolumn{1}{|C{1cm}|}{\boldmath{$A_u$}} &
  \multicolumn{1}{|C{1cm}|}{\boldmath{$A_s$}} &
  \multicolumn{1}{|C{1cm}|}{\boldmath{$h$}} &
  \multicolumn{1}{|C{1cm}|}{\boldmath{$A_u$}} &
  \multicolumn{1}{|C{1cm}|}{\boldmath{$A_s$}} &
  \multicolumn{1}{|C{1cm}|}{\boldmath{$h$}} &
  \multicolumn{1}{|C{1cm}|}{\boldmath{$A_u$}} &
  \multicolumn{1}{|C{1cm}|}{\boldmath{$A_s$}} &
  \multicolumn{1}{|C{1cm}|}{\boldmath{$h$}} \\
  \hline
  \hline
\multicolumn{10}{|l|}{\textbf{MFU Dataset}} \\ 
\hline
MFU 2000           & .74 & .99 & .85 & .92 & .93 & .92 & .92 & .99 & \textbf{.96} \\ 
\hline
\hline
\multicolumn{10}{|l|}{\textbf{GW Dataset}}\\
\hline
GW Split 1           & .01 & .96 & .03 & .24 & .95 & \textbf{.39} & .15 & .97 & .27 \\
\hline
GW Split 2           & .10 & .98 & .19 & .30 & .98 & .46 & .30 & .98 & \textbf{.46} \\
\hline
GW Split 3           & .09 & .97 & .17 & .40 & .94 & \textbf{.56} & .35 & .96 & .51 \\
\hline
GW Split 4           & .04 & .94 & .08 & .31 & .92 & \textbf{.47} & .25 & .95 & .39 \\
\hline
\hline
\multicolumn{10}{|l|}{\textbf{IAM Dataset}}\\
\hline
ZSL Split            & .58 & .88 & .70 & .71 & .82 & .76 & .77 & .93 & \textbf{.84} \\
\hline
Standard Split       & .46 & .87 & .61 & .64 & .82 & .72 & .70 & .90 & \textbf{.79} \\ \hline
\hline
\end{tabular}%
}
\caption{Generalized ZSL accuracy on various splits, with Cosine Similarity as the similarity metric $A_u$ = Accuracy with unseen word classes, $A_s$ = Accuracy with seen word classes, Generalized ZSL accuracy, $h$ = Harmonic mean of $A_u$ and $A_s$.}
\label{table:gzslresults}
\end{table*}

% euclidean distance 
\begin{table*}[h]
\centering
\resizebox{\textwidth}{!}{%
\begin{tabular}{|l|c|c|c|c|c|c|c|c|c|}
\hline
\hline
\multicolumn{1}{|c|}{\textbf{Split}} &
  \multicolumn{3}{|c|}{\textbf{PHOC}} &
  \multicolumn{3}{|c|}{\textbf{PHOS}} &
  \multicolumn{3}{|c|}{\textbf{Pho(SC)}} \\ 
  \hline
\multicolumn{1}{|c|}{} &
  \multicolumn{1}{|C{1cm}|}{\boldmath{$A_u$}} &
  \multicolumn{1}{|C{1cm}|}{\boldmath{$A_s$}} &
  \multicolumn{1}{|C{1cm}|}{\boldmath{$h$}} &
  \multicolumn{1}{|C{1cm}|}{\boldmath{$A_u$}} &
  \multicolumn{1}{|C{1cm}|}{\boldmath{$A_s$}} &
  \multicolumn{1}{|C{1cm}|}{\boldmath{$h$}} &
  \multicolumn{1}{|C{1cm}|}{\boldmath{$A_u$}} &
  \multicolumn{1}{|C{1cm}|}{\boldmath{$A_s$}} &
  \multicolumn{1}{|C{1cm}|}{\boldmath{$h$}} \\
  \hline
  \hline
\multicolumn{10}{|l|}{\textbf{MFU Dataset}} \\ 
\hline
MFU 2000           & .70 & 1 & .82 & .85 & .93 & \textbf{.89} & .8 & 1 & \textbf{.89} \\ 
\hline
\hline
\multicolumn{10}{|l|}{\textbf{GW Dataset}}\\
\hline
GW Split 1           & .01 & .97 & .02 & .23 & .93 & \textbf{.37} & .13 & .96 & .23 \\
\hline
GW Split 2           & .11 & .98 & .20 & .31 & .98 &  \textbf{.48} & .24 & .99 & .38 \\
\hline
GW Split 3           & .1 & .97 & .17 & .27 & .95 &  \textbf{.42} & .26 & .96 & .41 \\
\hline
GW Split 4           & .03 & .94 & .05 & .19 & .91 &  \textbf{.32} & .14 & .95 & .24 \\
\hline
\hline
\multicolumn{10}{|l|}{\textbf{IAM Dataset}}\\
\hline
ZSL Split            & .50 & .87 & .64 & .69 & .81 & .75 & .66 & .92 &  \textbf{.77} \\
\hline
Standard Split       & .40 & .87 & .54 & .60 & .81 &  \textbf{.69} & .54 & .90 & .68 \\ \hline
\hline
\end{tabular}%
}
\caption{Generalized ZSL accuracy on various splits with Euclidean Distance as the similarity metric. $A_u$ = Accuracy with unseen word classes, $A_s$ = Accuracy with seen word classes, Generalized ZSL accuracy, $h$ = Harmonic mean of $A_u$ and $A_s$.}
\label{table:gzslresultseuclid}
\end{table*}

\begin{table*}[h]
\centering
\resizebox{\textwidth}{!}{%
\begin{tabular}{|l|c|c|c|c|c|c|c|c|c|}
\hline
\hline
\multicolumn{1}{|c|}{\textbf{Split}} &
  \multicolumn{3}{|c|}{\textbf{Vanilla-CTC \color{black}}} &
  \multicolumn{3}{|c|}{\textbf{ PHO(SC)-CTC \color{black}}} &
  \multicolumn{3}{|c|}{\textbf{Pho(SC)NET}} \\ 
  \hline
\multicolumn{1}{|c|}{} &
  \multicolumn{1}{|C{1cm}|}{\boldmath{$A_u$}} &
  \multicolumn{1}{|C{1cm}|}{\boldmath{$A_s$}} &
  \multicolumn{1}{|C{1cm}|}{\boldmath{$h$}} &
  \multicolumn{1}{|C{1cm}|}{\boldmath{$A_u$}} &
  \multicolumn{1}{|C{1cm}|}{\boldmath{$A_s$}} &
  \multicolumn{1}{|C{1cm}|}{\boldmath{$h$}} &
  \multicolumn{1}{|C{1cm}|}{\boldmath{$A_u$}} &
  \multicolumn{1}{|C{1cm}|}{\boldmath{$A_s$}} &
  \multicolumn{1}{|C{1cm}|}{\boldmath{$h$}} \\
  \hline
  \hline
\multicolumn{10}{|l|}{\textbf{MFU Dataset}} \\ 
\hline
MFU 2000           & .99 & 1.0 & .99 & 1.0 & 1.0 & \textbf{1.0} & .92 & .99 & .96 \\ 
\hline
\hline
\multicolumn{10}{|l|}{\textbf{GW Dataset}}\\
\hline
GW Split 1           & .50 & .95 & \textbf{.65} & .50 & .91 & .64 & .15 & .97 & .27 \\
\hline
GW Split 2           & .63 & .95 & .76 & .75 & .96 & \textbf{.84} & .30 & .98 & .46 \\
\hline
GW Split 3           & .62 & .90 & .73 & .70 & .88 & \textbf{.78} & .35 & .96 & .51 \\
\hline
GW Split 4           & .44 & .91 & .59 & .60 & .89 & \textbf{.72} & .25 & .95 & .39 \\
\hline
\hline
\multicolumn{10}{|l|}{\textbf{IAM Dataset}}\\
\hline
ZSL Split            & .65 & .82 & .73 & .65 & .81 & .72 & .77 & .93 & \textbf{.84} \\
\hline
Standard Split       & .54 & .82 & .65 & .56 & .83 & .67 & .70 & .90 & \textbf{.79} \\ \hline
\hline
\end{tabular}%
}
\caption{Generalized ZSL accuracy on various splits. $A_u$ = Accuracy with unseen word classes, $A_s$ = Accuracy with seen word classes, $h$ = Harmonic mean of $A_u$ and $A_s$.}
\label{table:netvsctc}
\end{table*}

\begin{table*}[h]
\centering
\resizebox{12 cm}{!}{%
\begin{tabular}{|l|c|c|c|}
\hline
\hline
\multicolumn{1}{|c|}{\textbf{Split}} &
  \multicolumn{1}{|c|}{\textbf{ PHO(SC)-CTC \color{black}}} &
  \multicolumn{1}{|c|}{\textbf{ Pho(SC)NET \color{black}}} &
  \multicolumn{1}{|c|}{\textbf{AttentionHTR}} \\ 
  \hline
\multicolumn{1}{|c|}{} &
  \multicolumn{1}{|C{4cm}|}{\boldmath{$Accuracy$}}  &
  \multicolumn{1}{|C{4cm}|}{\boldmath{$Accuracy$}}  &
  \multicolumn{1}{|C{4cm}|}{\boldmath{$Accuracy$}} \\
  \hline
 \hline
\multicolumn{3}{|l|}{\textbf{GW Dataset}}\\
\hline
GW Split 1           & .91 & \textbf{.97} & .91 \\
\hline
GW Split 2           & .96 & \textbf{.98} & .94 \\
\hline
GW Split 3           & .88 & \textbf{.96} & .93 \\
\hline
GW Split 4           & .89 & \textbf{.95} & .88 \\
\hline
\hline
\multicolumn{3}{|l|}{\textbf{IAM Dataset}}\\
\hline
ZSL Split           & .81 & \textbf{.93} & .88 \\
\hline
Standard Split      & .83 & .90 & \textbf{.91} \\ 
\hline
\hline
\end{tabular}%
}
\caption{Comparision of PHO(SC)-CTC and PHo(SC)NET with AttentionHTR on seen classes.}
\label{table:attentionvsctc}
\end{table*}

\subsection{Training and Baselines}
The Pho(SC)Net was trained using an Adam optimizer with learning rate 1e-4, weight decay set as 5e-5, momentum at 0.9. The batch size is kept as 16. The hyper-parameters $\lambda_c$ and $\lambda_s$ were fine tuned using the validation set. The final values chosen for these parameters are 1 and 4.5 respectively. We also conducted ablation studies with the individual PHOCNet (SPP-PHOCNet) and the PHOSNet counter part. These two networks were also trained in a similar fashion for all the three datasets. This helps to investigate the effect of adding visual shape representations to the default attribute signature vector (PHOC vector). Early stopping was applied to the training process using reducedLR on plateau on the validation set. As this is the first work to perform zero-shot word recognition on publicly available benchmark datasets, we compare Pho(SC)Net with the classical SPP-PHOCNet model (trained using the settings in\cite{sudholtF16}).

\begin{figure*}[h]
\centering
\includegraphics[width=\textwidth, height=13cm]{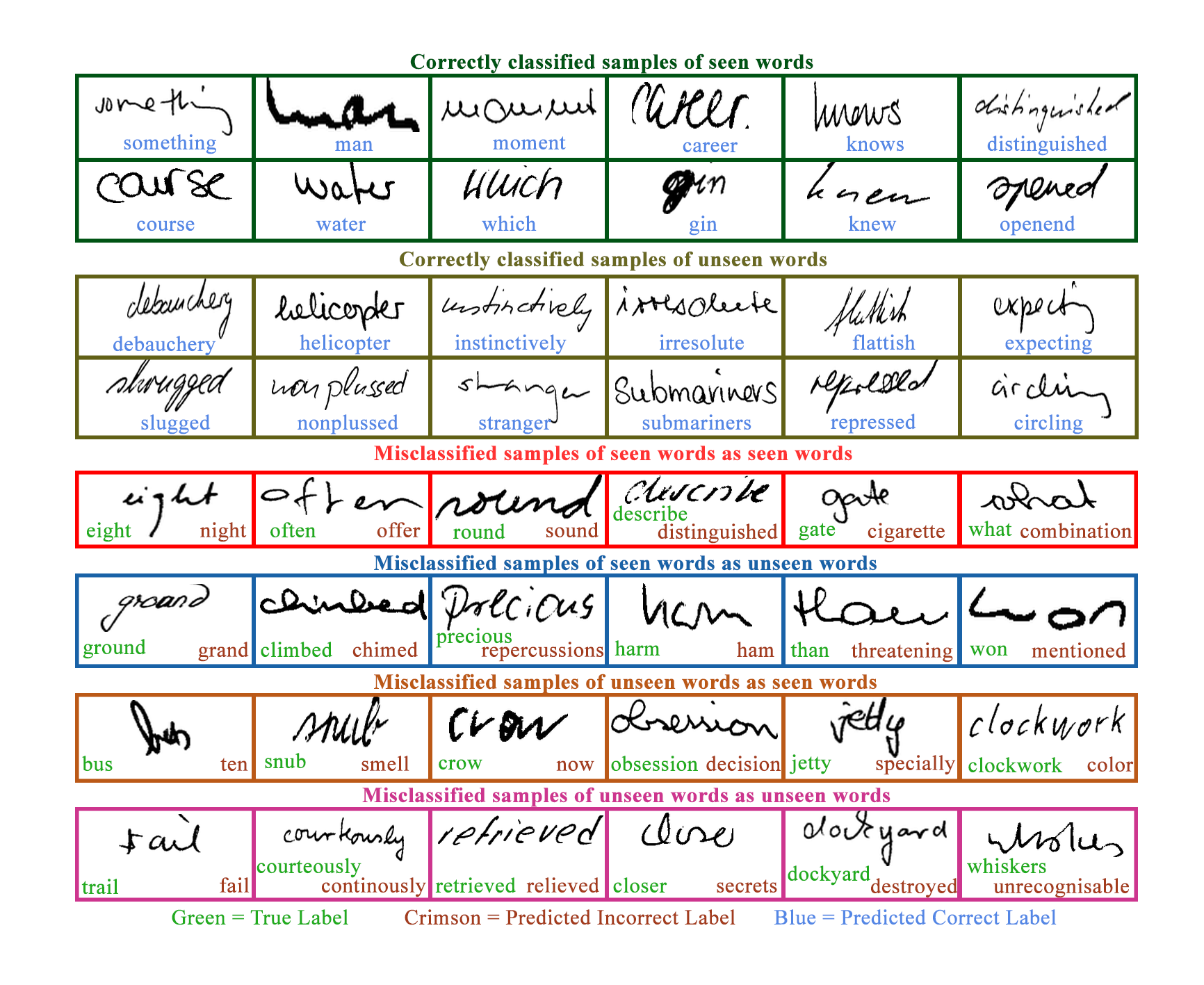}
\caption{Examples of correct and incorrect predictions in Generalized ZSL setting from the standard IAM Split}
 \label{fig:predictions}
\end{figure*}

\begin{figure*}[h]
\centering
\includegraphics[width=12cm,height=8cm]{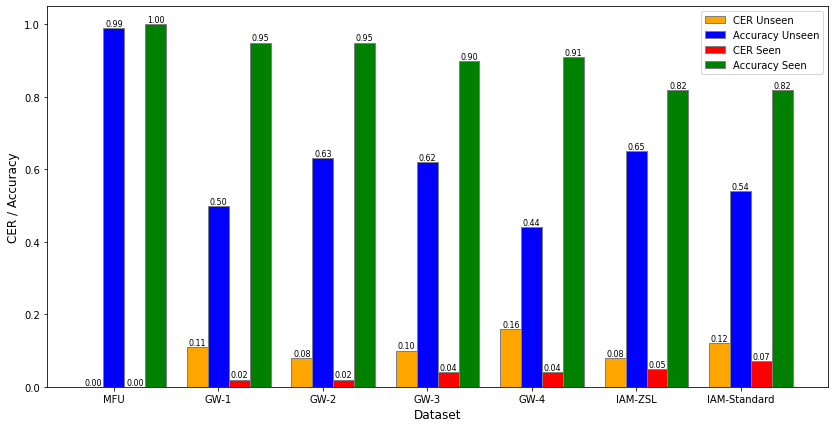}
\caption{Generalized ZSL accuracy and Character Error Rate (CER) of Vanilla-CTC \color{black} on various datasets.}
 \label{fig:CTCbar}
\end{figure*}

The Pho(SC)-CTC and Vanilla-CTC \color{black} models were trained using an Adam optimizer with a learning rate of 0.001. Early stopping was applied on the validation set to avoid over-fitting. We compare Pho(SC)-CTC with Vanilla-CTC \color{black} and Pho(SC)net.\newline
There is no defined search space for a CTC framework as the network's output is a sequence of characters. So we cannot compare it with the conventional zeroshot learning results of Pho(SC)net. Thus we compare both the models in the generalized setting.
\footnote{\url{https://github.com/raviRB/Pho-SC--CTC}}.

\subsection{Performance Metrics}
The top-1 accuracy of the model's prediction is used as the performance metric for all the experiments. The top-1 accuracy measures the proportion of test instances whose predicted attribute signature vector is closest to the true attribute signature vector. At test time, in the ZSL setting, the aim is to assign an unseen class label, i.e. $\mathcal{Y}^u$ to the test word image and in the generalized ZSL setting (GZSL), the search space includes both seen and unseen word labels i.e. $\mathcal{Y}^u \cup \mathcal{Y}^s$. Therefore, in the ZSL setting, we estimate the top-1 accuracy over $\mathcal{Y}^u$. In the GZSL setting, we determine the top-1 accuracy for both $\mathcal{Y}^u $ and $ \mathcal{Y}^s$ independently and then compute their harmonic mean. \newline
The output of CTC is a sequence of characters, thus we can calculate the character error rate (CER) for a set of testing images T, using the following formula.
\begin{linenomath}
\begin{equation}
    CER = \sum_{i=1}^{T} \frac{ED(y_{ip},y_i)}{\lvert y_i \rvert}
\end{equation}
\end{linenomath}
here $y_{ip}$ is the predicted label for the input image $x_i$ and $y_i$ is the true label, ED(a,b) is the minimum number of insertion,deletion and substitutions required to change a to b, i.e the edit distance between a and b.
\begin{figure*}[h]
\centering
\includegraphics[width=12cm,height=8cm]{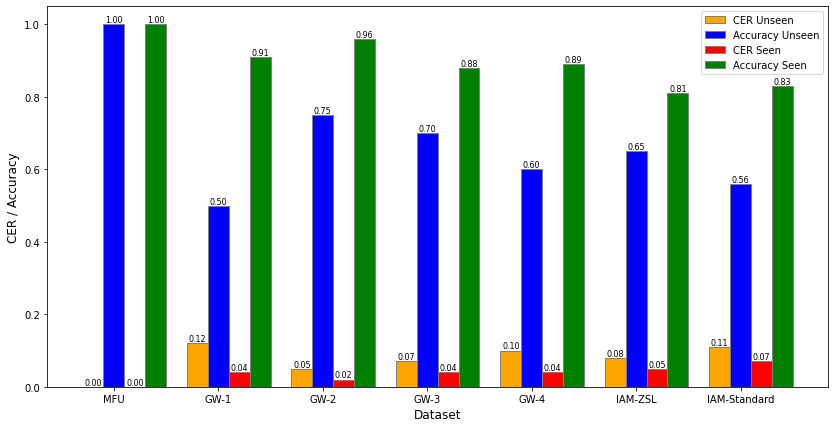}
\caption{Generalized ZSL accuracy and Character Error Rate (CER) of Pho(SC)-CTC \color{black} on various datasets.}
 \label{fig:CTC-p-bar}
\end{figure*}

\section{Results and Discussion}
It is evident from the results in Tables [\ref{table:convzslresult},\ref{table:gzslresults},\ref{table:gzslresultseuclid}] that cosine similarity performed better than Euclidean distance as the similarity matric; thus, the discussion is based on the results of using cosine similarity as the similarity matric\color{black}. The accuracies for the unseen word labels under the conventional ZSL setting is presented in Table \ref{table:convzslresult}. Overall, it is observed that the PHOC representation is not well-suited for predicting unseen word labels. However, the PHOS representation is more accurate in predicting the unseen word labels. Further, the combination of both the vectors (Pho(SC)) results in a significant improvement (on an average $>5\%$) in the unseen word prediction accuracy. The MFU dataset, on the account of being synthetically generated and noise-free, has the highest unseen word recognition accuracy. We obtained the least accuracy on the GW dataset split 4. We attribute this low accuracy to a rather large number of unseen word classes in the test set for this split.

It is also observed that the accuracy of the model on MFU and IAM datasets are significantly higher than that of any of the GW splits. This is explained by observing the number of seen classes the model is presented during training. Both MFU and IAM datasets have more than 2000 seen word labels, allowing the model to learn the rich relationships between the word labels as encoded by the attribute signatures. Learning this relationship is essential for the model to perform well on unseen word images.

We also observe that using only PHOS as the attribute signature results in better performance than using PHOC on datasets that have homogeneous writing style. This is inferred by noticing the significant increase in the unseen word accuracy of over 14\% on the GW dataset splits by PHOS over PHOC.

Table \ref{table:gzslresults} presents the test seen and unseen word accuracies, along with the harmonic mean for the GZSL setting. The high accuracies on the seen word labels and low accuracies on unseen word labels for the model using only PHOC seems to suggest that the PHOC representation is more suitable for scenarios where the word labels across the train and test are similar. In contrast, the PHOS model has marginally lower accuracies on the test seen word labels (in comparison to PHOC), but significantly higher accuracies on the unseen word labels (again in comparison to PHOC). This indicates that looking at visual shapes is more reliable when the train and test sets contain different word labels. Further, we also observe that the combined model yields higher performance for both MFU and IAM datasets that have a large number of seen classes. However, on the GW dataset, that has a significantly smaller number of seen word classes, the PHOS model is able to even outperform the combined model. The poor performance of the PHOC model that is pulling down the performance of the combined model on the unseen classes for the GW dataset can also be attributed to the small number of seen word classes that the model is exposed to during training. However, note that the PHOC model achieves significantly higher accuracies on the test images of the seen word classes, indicating that the model has not overfit, but is biased towards these classes. Table  \ref{table:netvsctc} presents the results for the various CTC models considered in this work.\newline 
Vanilla-CTC \color{black} performs better than Pho(SC)net on the GW dataset and MFU dataset, but for the IAM dataset Pho(SC)Net outperforms Vanilla-CTC \color{black} by a considerable margin. Pho(SC)-CTC \color{black} performs better than Vanilla-CTC \color{black} on GW Split 2, GW Split 3, GW Split 4, and IAM dataset in standard setting. In other datasets, both the models show similar performance. We observe that Pho(SC)-CTC \color{black} performs better when there is not much heterogeneity in the dataset, furthermore Pho(SC)Net is able to capture some rich features from the images, which are not being extracted by Vanilla-CTC \color{black}.
\subsection{Comparison with State of the Art Model for Seen Classes}
Phos(SC)-CTC has been compared with AttentionHTR \cite{attentionHTR} which is the current state-of-the-art model for seen handwritten word recognition. AttentionHTR \cite{attentionHTR} is an attention-based sequence to sequence model consisting of four stages:(i)Transformation stage for normalizing the input;(ii)feature extraction stage;(iii)sequence modeling stage, and (iv) Prediction stage. The authors in  \cite{attentionHTR} have suggested transfer learning for the IAM dataset. The model for the GW dataset was trained from scratch. Table \ref{table:attentionvsctc} shows the comparison of seen class accuracy between Pho(SC)net, Pho(SC)-CTC, and AttentionHTR. Pho(SC)Net outperforms AttentionHTR on all other splits except one. \color{black}

\subsection{Error Analysis}
Figure \ref{fig:predictions} illustrates a few predictions by the Pho(SC) model on the IAM dataset under the GZSL setting. It can observed that even when the model incorrectly predicts the word, there is a good overlap between the characters of the predicted and true word label.\newline
\Cref{fig:CTC-p-bar,fig:CTCbar} shows the CER for all the datasets with the word accuracies. It can be observed that the CER is quite less even when the word accuracies are high, signifying that there is a good overlap between the characters of the predicted and true word label.

\begin{figure}[h]
% \centering
\includegraphics [width=\linewidth]{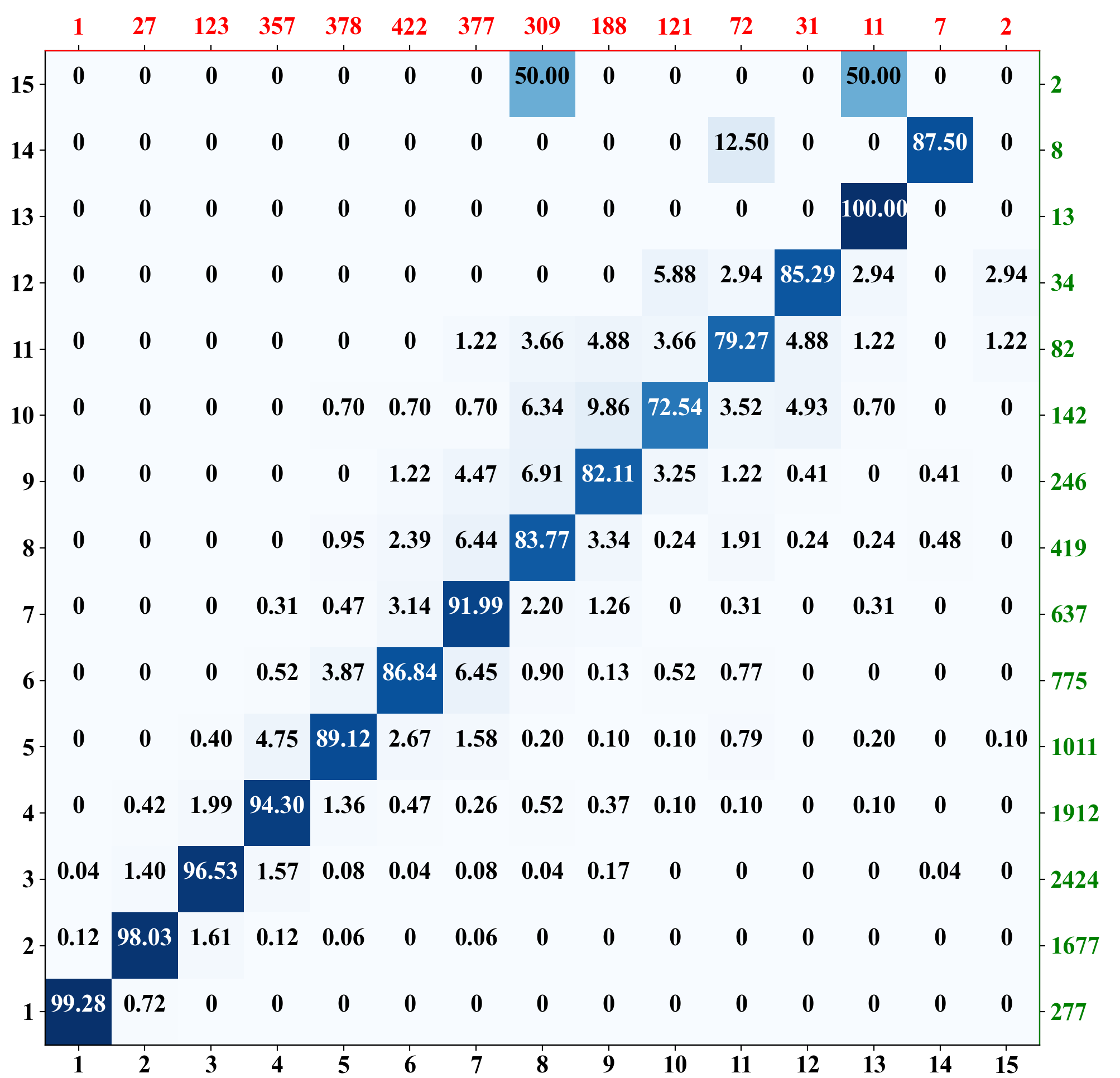}
\caption{Lengthwise confusion matrix(normalised) for predicted word length(left axis) and true word length(bottom) for  Standard IAM Split. Labels on top(red) indicate the number of word classes in lexicon of that length while on right(green) represent the number of samples in test set for the corresponding word length.}
 \label{fig:confmat}
\end{figure}

Figure \ref{fig:confmat} presents the confusion matrix for predictions on the IAM dataset standard split in the GZSL setting. The confusion matrix has been computed between words of different lengths to uncover any biases of the model (if any). It is difficult to visualize the class specific confusion matrix as there are over 1000 word labels with very few (often only 1) images per word label. The length of the predicted word labels is mostly within a range of the length of the true word label. In general, the model is not biased towards words of any specific length. The high values along the diagonal indicates that the model is often predicting the word of the correct length (except when the word length is 15, for which there are only 2 samples and 2 classes).

\section{Conclusion}
We propose a hybrid model that combines novel attribute signature representation (PHOSC) (that characterizes the occurrence and position of elementary visual shapes  and characters in a word) along with LSTM in a CTC framework. Our experiments demonstrate the effectiveness of PHOSC-CTC for predicting unseen word labels in the generalized ZSL setting for two datasets with more homogeneous handwriting, while the classical Pho(SC)Net representation is more suitable for recognition of unseen word classes with more heterogeneous handwriting. Future directions to this work include extending the PHOS representations to include other non-Latin scripts like Arabic, Chinese, and Indic scripts.

\bibliographystyle{sn-basic}
\bibliography{bibliography}% common bib file


\end{document}